%% file: chatgraph.tex
\documentclass[conference]{IEEEtran}
\IEEEoverridecommandlockouts
\usepackage{cite}
\usepackage{amsmath,amssymb,amsfonts}
\usepackage{algorithmic}
\usepackage{graphicx}
\usepackage{textcomp}
\usepackage{xcolor}

\def\BibTeX{{\rm B\kern-.05em{\sc i\kern-.025em b}\kern-.08em
    T\kern-.1667em\lower.7ex\hbox{E}\kern-.125emX}}
\begin{document}

\title{ChatGraph: Chat with Your Graphs}

\author{\IEEEauthorblockN{Yun Peng$^1$, Sen Lin$^1$, Qian Chen$^2$, Lyu Xu$^2$, Xiaojun Ren$^1$, Yafei Li$^3$, Jianliang Xu$^2$}
\IEEEauthorblockA{
\textit{$^1$Guangzhou University, $^2$Hong Kong Baptist University, $^3$Zhengzhou University}\\
\{yunpeng,slin,xjren\}@gzhu.edu.cn, \{qchen,cslyuxu,xujl\}@comp.hkbu.edu.hk, ieyfli@zzu.edu.cn}
}

\maketitle
\begin{abstract}
Graph analysis is fundamental in real-world applications. Traditional approaches rely on SPARQL-like languages or clicking-and-dragging interfaces to interact with graph data. However, these methods either require users to possess high programming skills or support only a limited range of graph analysis functionalities. To address the limitations, we propose a large language model (LLM)-based framework called ChatGraph. With ChatGraph, users can interact with graphs through natural language, making it easier to use and more flexible than traditional approaches. {\it The core of ChatGraph lies in generating chains of graph analysis APIs based on the understanding of the texts and graphs inputted in the user prompts.} To achieve this, ChatGraph consists of three main modules: an API retrieval module that searches for relevant APIs, a graph-aware LLM module that enables the LLM to comprehend graphs, and an API chain-oriented finetuning module that guides the LLM in generating API chains. 
We have implemented ChatGraph and will showcase its usability and efficiency in four scenarios using real-world graphs. 
\end{abstract}


\newtheorem{definition}{Definition}
\newtheorem{example}{Example}
\newtheorem{theorem}{Theorem}
\newtheorem{lemma}{Lemma}
\newtheorem{proposition}{Proposition}
\newtheorem{corollary}{Corollary}

\input{sec-introV2}
\input{sec-backV2}

\input{sec-scenario}
\input{sec-conc}



\bibliographystyle{IEEEtran}
\bibliography{IEEEabrv,newpg}

\end{document}

%% file: sec-introV2.tex
\section{Introduction}

Graphs have been widely used to model unstructured data in various real-world applications, such as chemi-informatics, bioinformatics, and computer vision, among others. Traditional approaches to interact with graph data include the declarative language ({\it e.g.}, SPARQL \cite{10.1145/1567274.1567278} and GraphQL \cite{GQL}) and the clicking-and-dragging interface ({\it e.g.}, GBLENDER \cite{GBLENDER} and CATAPULT \cite{CATAPULT}). However, the former approach requires users to possess high programming skills in order to accurately express their needs, which may not be practical for domain experts who are not proficient in programming. On the other hand, the latter approach currently supports only a limited range of graph processing functionalities and requires significant effort to extend beyond its current capabilities.

Recently, there has been a growing trend in the use of large language model (LLM)-based frameworks for data interaction. Examples of such frameworks include ChatGPT \cite{brown2020language}, LangChain \cite{yao2023react}, Semantic Kernel \cite{SemanticKernel}, and Gorilla \cite{patil2023gorilla}. 
The third-party data analysis APIs can be integrated by LLMs. Users can specify their needs using natural language, and the LLM recognizes the user's intent and invokes an appropriate chain of APIs. Data interaction using LLM-based frameworks becomes significantly easier to use and more flexible compared to traditional approaches.

Unfortunately, existing systems do not support graphs.
In graph analysis tasks, users often need to provide both textual information and graphs in their prompts. For LLMs to effectively handle these prompts, they must possess a certain level of understanding of the graphs. Existing methods that support multi-modal prompts can be broadly classified into two categories. The first is designing multi-modal LLMs ({\it e.g.} GPT-4V \cite{yang2023dawn} and Next-GPT \cite{nextgpt}). The second is finetuning uni-modal LLMs with multi-modal instructions ({\it e.g.}, Gorilla \cite{patil2023gorilla}). However, despite the consideration of text, image, and voice in existing works, graphs have not been thoroughly studied.

To address the aforementioned limitations, this demonstration proposes a novel framework called ChatGraph, which falls into the second category of existing approaches. ChatGraph incorporates three key ideas to enhance the capabilities of LLMs in graph analysis. First, we develop a graph-aware LLM module that enables the LLM to understand graphs. Second, we introduce an API chain-oriented finetuning module that guides the LLM in generating API chains. Third, 
we propose an API retrieval module that searches for relevant APIs in the high-dimensional space, which is critical for performance \cite{patil2023gorilla}.

\noindent
{\bf Contributions.}
The contributions of this work are as follows.

\begin{itemize}
\item We design and implement ChatGraph, an LLM-based framework to interact with graphs. To the best of our knowledge, this is the first chat-based LLM framework to interact with graphs.
\item We provide three powerful modules: API retrieval, graph-aware LLM, and API chain-oriented finetuning to support graph analysis through natural language. 
\item We demonstrate how our ChatGraph works in four real-world scenarios using diverse graph datasets.
\end{itemize}

\noindent
{\bf Organization.} In the remainder of this demonstration paper, we will provide an overview of ChatGraph in Sec.~\ref{sec:back}. Following that, we present the implementation details and demonstration scenarios in Sec.~\ref{sec:impl} and Sec.~\ref{sec:demo}, respectively.

\vspace{-1ex}
\begin{figure}[tbp] 
\centering
\includegraphics[width=9cm]{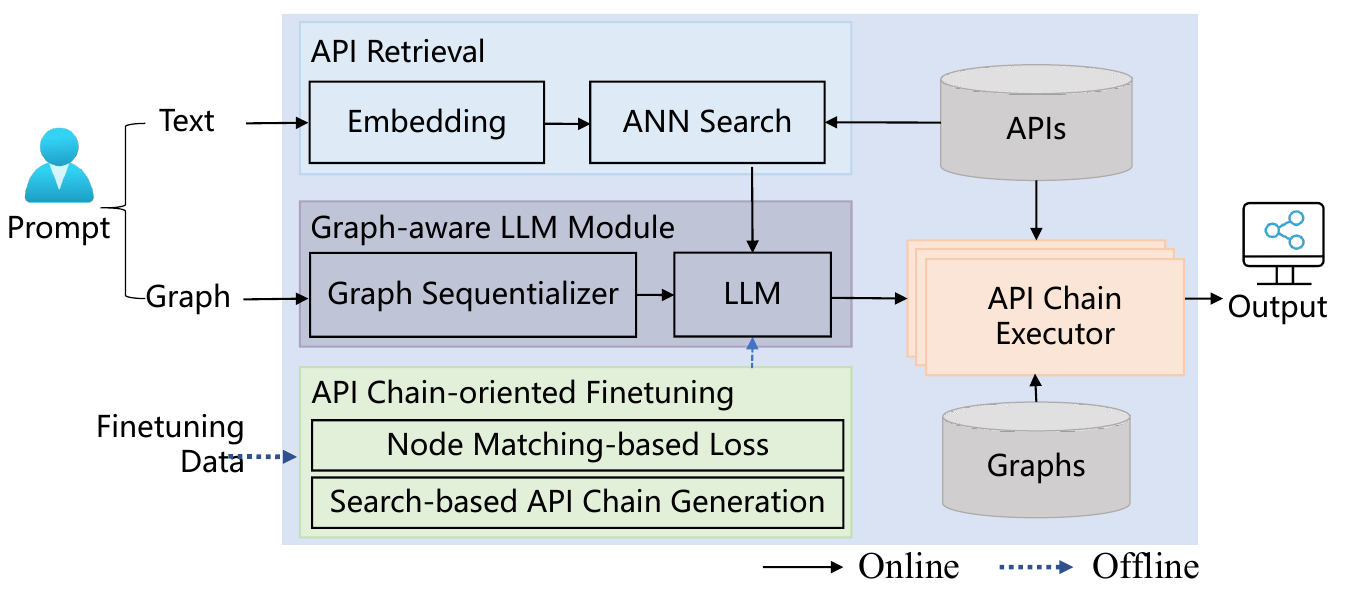}
\vspace{-5ex}
\caption{System overview of ChatGraph}\label{fig:overview}
\vspace{-2ex}
\end{figure}

%% file: sec-backV2.tex
\section{ChatGraph Overview}\label{sec:back}

A graph $G$ is a tuple $(V,E)$, where $V$ is the set of nodes and $E$ is the set of edges. An API chain is denoted by $C$ and an API is denoted by $a$.

\subsection{System Overview}

Fig.~\ref{fig:overview} presents the architecture of ChatGraph. Suppose a user inputs a text and a graph in her prompt. The API retrieval module consists of an embedding sub-module and an ANN search sub-module. The text of the prompt is first embedded into a vector, and then the APIs whose embeddings are the most similar vectors to the text's embedding vector are found. The graph-aware LLM module consists of a graph sequentializer sub-module and an LLM sub-module. The graph of the prompt is first transformed into sequences and the LLM  generates the API chain. The generation is guided by the API chain-oriented finetuning module, which consists of a node matching-based loss sub-module and a search-based prediction sub-module. 

The (sub)-modules, including the graph sequentializer, the API chain-oriented finetuning, and the ANN search, are presented below. Other (sub)-modules utilize existing methods.


\subsection{Graph Sequentializer}

Since the LLM requires sequential input data, we decompose a graph $G$ into paths.  A naive method is to enumerate all possible paths in $G$. However, since $G$ can have an exponential number of paths, we adopt the length-constrained path cover method proposed in our prior works \cite{LyuICDE,LyuSigmod}. The main idea is that for each node $u$ in $G$, we use the paths starting at $u$ of length at most $l$ to cover the subgraph of $G$ within $l$ hops of $u$. The total number of paths to cover $G$ is at most $O(|G|2^l)$, where $|G|$ is the number of nodes in $G$.

Considering that graphs often have multi-level structures ({\it e.g.}, the tertiary structure of proteins and the community structure of social networks), the graph sequentializer can produce paths at different levels. In particular, for a graph $G$, we adopt the method in \cite{RUM2019} to compute a super-graph for $G$, where each super-node represents a motif of $G$. The paths 
of $G$ and the super-graph of $G$ are inputted to the LLM to learn multi-level structures of $G$.

\subsection{API Chain-oriented Finetuning}

The API chain has two inherent properties. First, one node in the generated API chain should match one node in the ground-truth API chain. Second, there may be several API chains that are equivalent to answering the user's question. 
Therefore, we design two sub-modules in the API chain-oriented finetuning. The node matching-based loss sub-module is to encode the first property. The search-based prediction sub-module is to encode the second property.

{\bf Node matching-based loss function.}
The node matching-based loss function is widely used for graph matching \cite{GEDIJCAI}. The key ideas are to i) minimize the graph edit distance between the generated API chain and the ground-truth and ii) encode the one-to-one matching property.

Suppose $C$ and $C'$ denote the generated and the ground-truth API chains, respectively. Let $M$ denote the matching between the nodes of $C$ and $C'$, where $M_{i,j}=1$, if the node $u_i$ of $C$ matches the node $v_j$ of $C'$; otherwise, $M_{i,j}=0$.

\begin{definition}
The node matching-based loss function is  
$\min_M  X  + \alpha {Y}$ , where $X$ is the graph edit distance between $C$ and $C'$, $Y = \sum_{u_i\in C} (1-\sum_{v_k\in C'} M_{i,k})^2 + \sum_{v_k\in C'} (1-\sum_{u_i\in C} M_{i,k})^2$ is the regularizer to encode the one-to-one matching, and $\alpha$ is a weight to balance $X$ and $Y$. 
\end{definition}

\vspace{0ex}
{\bf Search-based prediction.} The API chain generation is to iteratively extend a partial chain to a full chain. In each iteration, an API is added. Since there are a large number of APIs, we adopt the search method with random rollouts to guide the prediction \cite{GEDIJCAI} to reduce the space of prediction. 

Let $C_p$ denote the partial API chain generated so far and $S$ denote the set of candidate APIs that can be used to extend $C_p$ in the current iteration. For each API $a$ in $S$, we conduct $r$ random rollouts. In each rollout, we randomly extend $C_p+\{a\}$ to a full chain $C$ and the loss between $C$ and a ground-truth API chain is used to score $a$. Since there may be several ground-truth API chains, the smallest loss of $C$ is used to score $a$. The API having the highest score is added to $C_p$. We repeat this procedure until a full chain is generated.

{\bf Dataset preparation.} 
The dataset for finetuning the LLM is prepared in this way. 
We recruited a group of students from the College of Chemistry to search for questions related to chemical molecules. Then, the students solved their 
questions by invoking APIs manually, and their actions were logged. Subsequently, we extracted the API chains from the logs as ground truths for the corresponding questions. 
We can also crawl questions and answers from chemistry software websites for finetuning data. 


\vspace{-1.1ex}
\subsection{Approximate Nearest Neighbor Search}

The descriptions of APIs and the text of user's prompt are embedded into high-dimensional vectors and ANN search in the embedding space finds the best API w.r.t user's prompt.

\begin{definition}
Given a set $D$ of data vectors, a query vector $h$, and a parameter $\epsilon$, the approximate nearest neighbor (ANN) search returns a vector $h'$ in $D$ s.t. $d(h',h)<(1+\epsilon)d(h^*, h)$, where $d$ is the distance function between two vectors and $h^*$ denotes the exact nearest neighbor of $h$ in $D$.
\end{definition}

Recent works \cite{ISsurvey, annsExpt, CompSurveyPGVLDB21} show that the proximity graph (PG) index outperforms other indexes on large-scale ANN search applications. A PG of $D$ is a graph, where each node is a vector in $D$ and two nodes have an edge if the two nodes satisfy some proximity property. Vectors that are similar to the query vector $h$
can be found through a greedy routing process within the PG. During each routing step, the router
navigates greedily to the neighbor that is closest to $h$. $\tau$-MG, proposed in our prior work \cite{tMG}, is considered the state-of-the-art PG. The key component of $\tau$-MG is its edge occlusion rule.

\begin{definition}
(Edge occlusion rule \cite{tMG}) Given three nodes $u$, $u'$, and $v$ in a $\tau$-MG, if the edge $(u,u')$ is in the $\tau$-MG and $u'$ is in the intersection of $ball(u,\delta(u,v))$ and $ball(v,\delta(u,v)-3\tau)$, then the edge $(u,v)$ is not in the $\tau$-MG, where $ball(u,r)$ denotes the ball centered at $u$ with radius $r$. 
\end{definition}

The 
nearest neighbor of $h$ can be found by a greedy routing on $\tau$-MG in  $O(n^{\frac{1}{m}}(\ln n)^2)$ time, which is smaller than all existing PGs.

%% file: sec-scenario.tex
\section{Implementation}\label{sec:impl}

ChatGraph is implemented based on the open source library Gradio.\footnote{https://www.gradio.app} The LLMs integrated are ChatGLM, MOSS, and Vicuna, which are downloaded from HuggingFace. The codes for the ANN search, the graph sequentializer, the node matching-based loss function, and the search-based prediction are modified from the source codes of the respective papers, which are obtained from GitHub.

\begin{figure}[t] 
\vspace{-2ex}
\centering
\includegraphics[width=9cm]{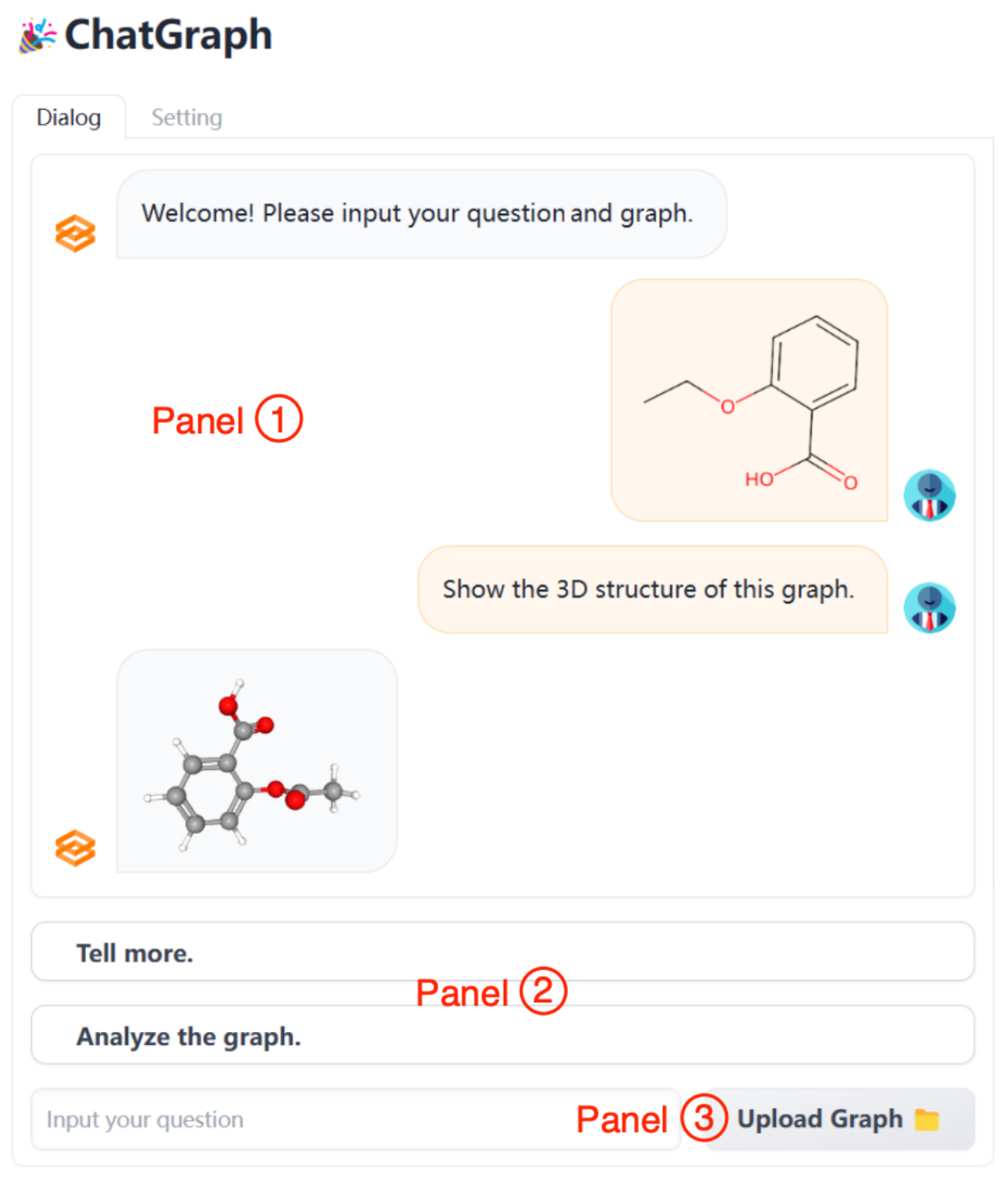}
\caption{Interface of ChatGraph}\label{fig:mainUI}
\vspace{-0ex}
\end{figure}


The interface of ChatGraph is shown in Fig.~\ref{fig:mainUI}, which consists of three panels. 
The dialogs between the user and the model is displayed in Panel \normalsize{\textcircled{\scriptsize{1}}}\normalsize. Panel \normalsize{\textcircled{\scriptsize{2}}}~\normalsize shows the suggested questions. The user can input her questions and upload graphs in Panel \normalsize{\textcircled{\scriptsize{3}}}\normalsize.

\begin{figure}[tp] 
\vspace{-2ex}
\centering
\includegraphics[width=9cm]{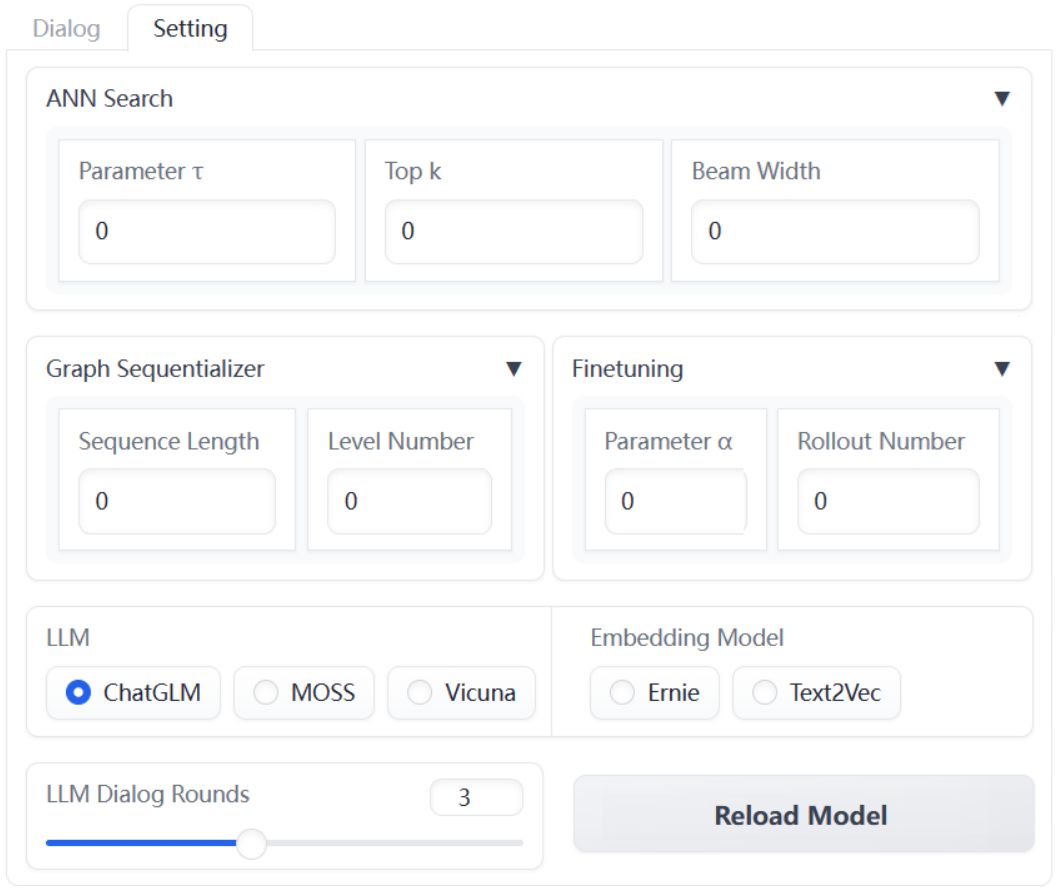}
\vspace{-0.2in}
\caption{Configuration parameters of ChatGraph}\label{fig:setting}
\vspace{-0ex}
\end{figure}

Fig.~\ref{fig:setting} shows the configuration parameters of ChatGraph. The left part of Fig.~\ref{fig:setting} shows the parameters for the ANN search, the graph sequentializer, and the finetuning modules. The right part of Fig.~\ref{fig:setting} shows the parameters for the LLM. 

\section{Demonstration Scenarios}\label{sec:demo}


In the demonstration, we will showcase the usability and efficiency of ChatGraph in four different scenarios. 

\subsubsection{Chat-based Graph Understanding}

\begin{figure}[tbp] 
\centering
\includegraphics[width=9cm]{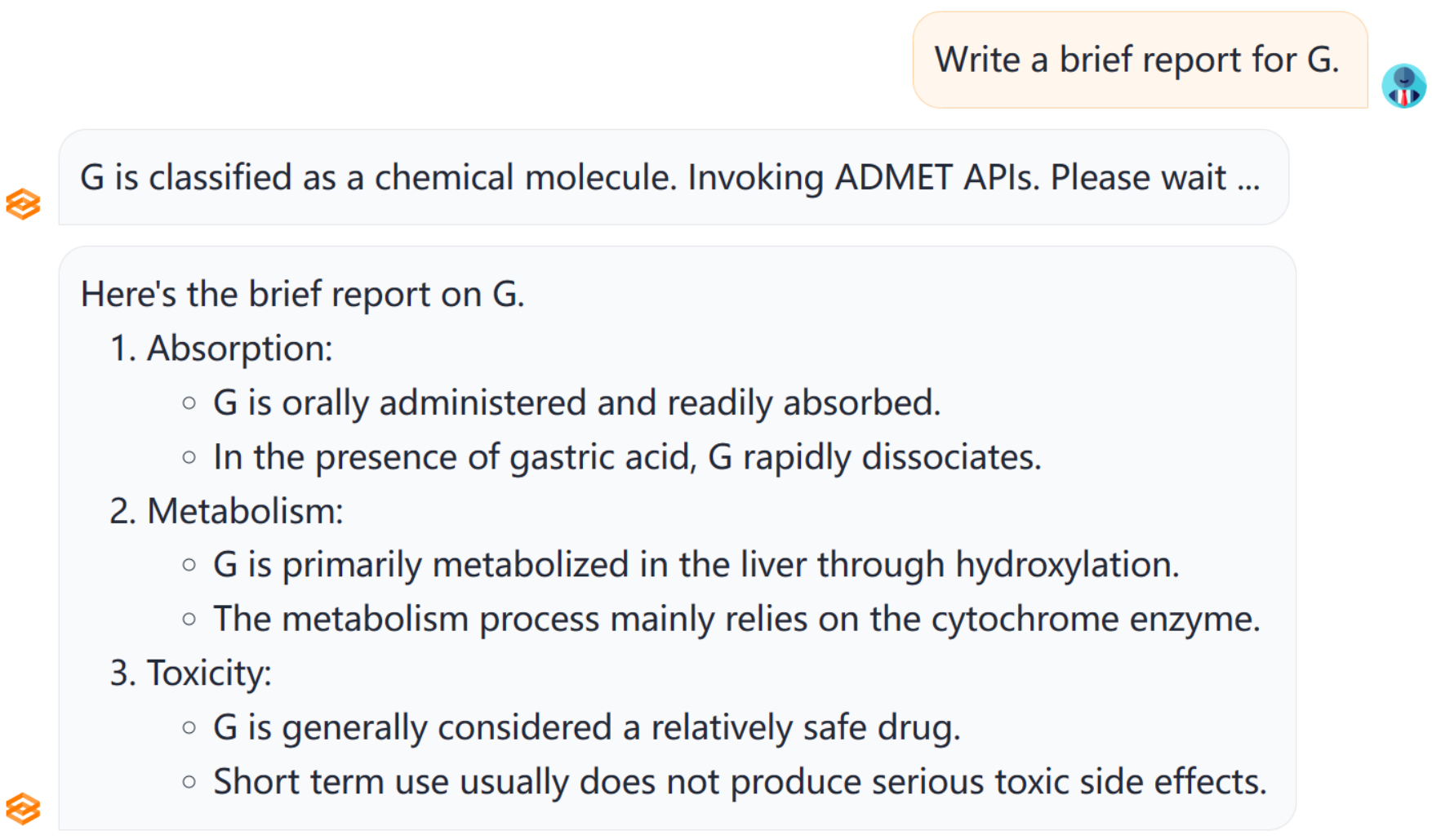}
\vspace{-0.2in}
\caption{Chat-based graph understanding}\label{fig:Gunderstand}
\vspace{-0ex}
\end{figure}

Graph understanding is important in real-world applications. For a social network, users are often interested in the social properties of the network. For a chemical molecule, users are often interested in its chemical properties. 
ChatGraph supports graph understanding through natural language. 
For example, as shown in Fig.~\ref{fig:Gunderstand}, a user submits a graph $G$ and a text ``Write a brief report for $G$'' in her prompt. ChatGraph first predicts the type of $G$. If $G$ is a social network, social-specific APIs ({\it e.g.}, community and connectivity) will be invoked to analyze $G$. Similarly, if $G$ is a molecule graph, molecule-specific APIs ({\it e.g.}, toxicity and solubility) will be invoked. A report is generated based on the results of the APIs.

\vspace{-0ex}
\vspace{-0in}

\subsubsection{Chat-based Graph Comparison}

\begin{figure}[tp] 
\centering
\includegraphics[width=9cm]{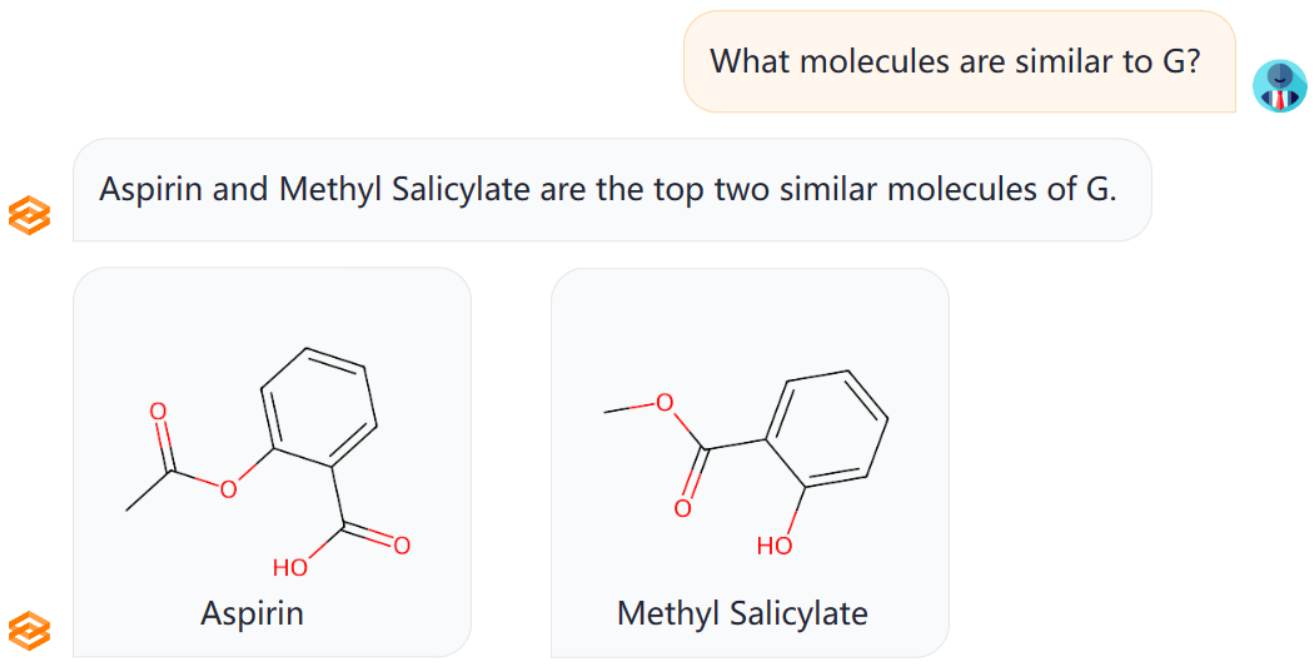}
\caption{Chat-based graph comparison}\label{fig:GCompare}
\vspace{-0ex}
\end{figure}

Graph comparison is a fundamental task in graph analysis. Drug designers use graph comparison in the virtual filtering of drug development. HRs use graph comparison in social networks to find the teams of their interest.
This scenario demonstrates the chat-based graph comparison. For example, as shown in Fig.~\ref{fig:GCompare}, a user submits a graph $G$ and a text ``What molecules are similar to $G$''. ChatGraph invokes the similarity search API for $G$ against a molecule graph database and outputs the top two similar molecules.


\subsubsection{Chat-based Graph Cleaning}

\vspace{-0ex}

\vspace{-0ex}
\begin{figure}[thbp] 
\centering
\includegraphics[width=8.5cm]{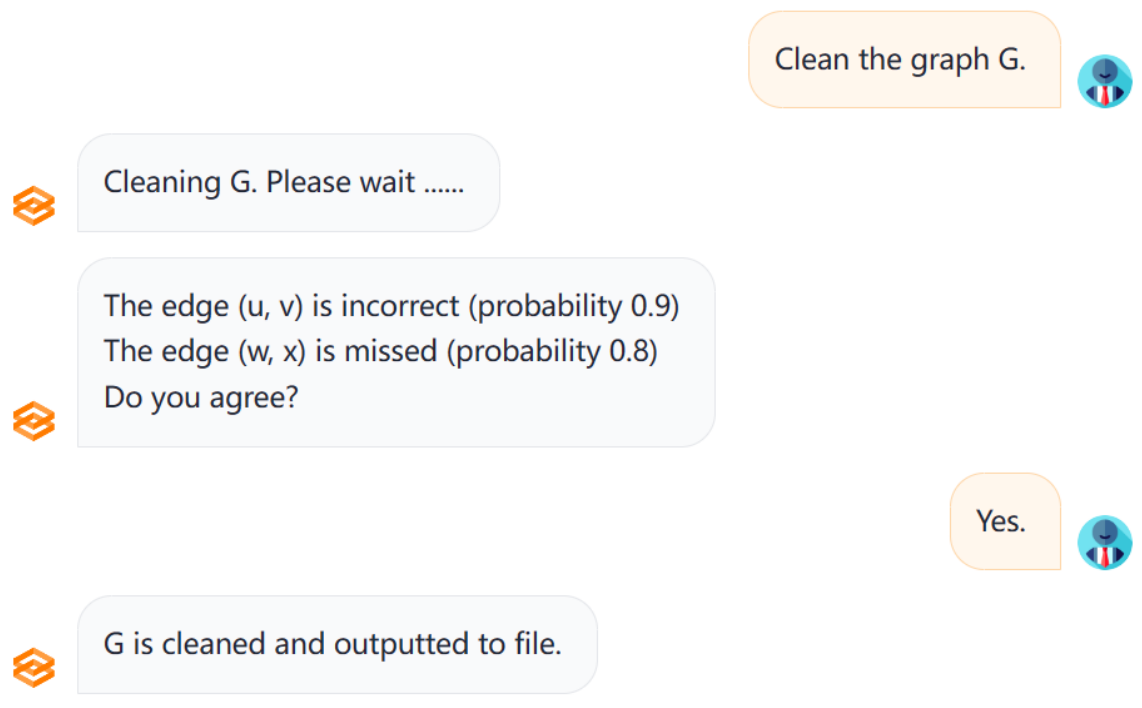}
\caption{Chat-based graph cleaning}\label{fig:GClean}
\vspace{-1ex}
\end{figure}

Graph cleaning is also important in practice. Data scientists need to clean the mislabels of graphs before using the graphs for model training. Users of knowledge graphs wish to reduce the noises in the knowledge graphs. 
We demonstrate the chat-based graph cleaning in this scenario. For example, as shown in Fig.~\ref{fig:GClean}, a user submits a knowledge graph $G$ and a text ``Clean $G$''. ChatGraph first invokes the knowledge inference APIs to detect the incorrect edges and the missing edges in $G$ and asks the user for confirmation. After that, the graph edit APIs are invoked to edit the edges in $G$.


\subsubsection{Chat-based API Chain Monitoring}

\begin{figure}[tp] 
\centering
\includegraphics[width=9cm]{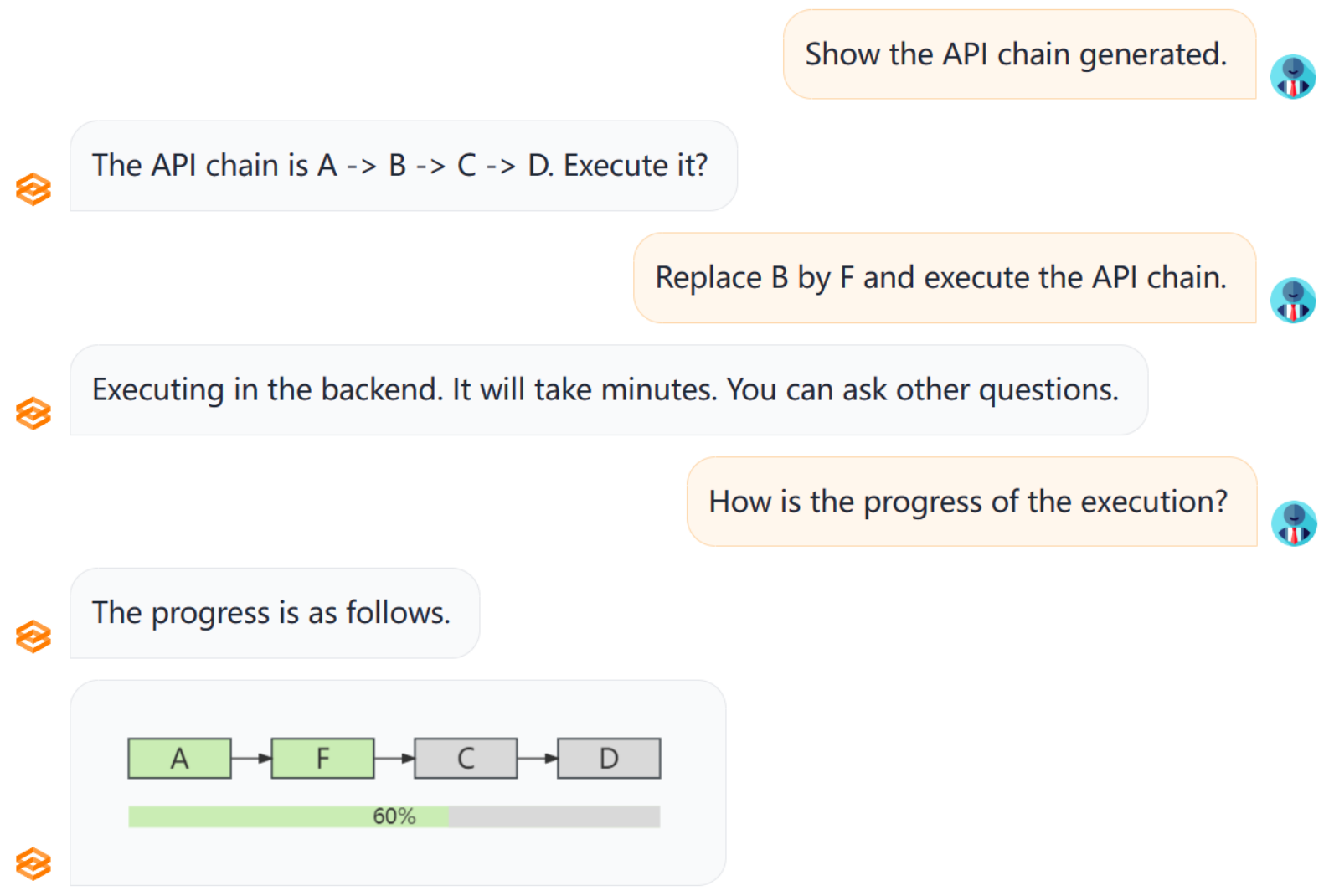}
\caption{Chat-based API chain monitoring}\label{fig:monitor}
\vspace{-0ex}
\end{figure}

The API chain generated by the LLM may not be completely correct. Therefore, users need to confirm the API chain before it is executed and edit it if needed. What is more, users may also wish to monitor the progress during the execution of the API chain.  An example of such monitoring is shown in Fig.~\ref{fig:monitor}.

\vspace{-0ex}

%% file: sec-conc.tex
\section{Conclusion}\label{sec:conc}

We have presented ChatGraph, a chat-based framework to interact with graphs. It is mainly composed of three modules: the API retrieval module, which adopts the SOTA $\tau$-MG method to search for APIs in the embedding space, the graph-aware LLM module, where a graph sequentializer is proposed to transform graphs into paths, and the API chain-oriented finetuning module, which consists of the node matching-based loss function and the search-based prediction. Furthermore, we demonstrate four scenarios using real-world graphs to showcase the usability and efficiency of ChatGraph.